\documentclass[letterpaper, 10 pt, conference]{ieeeconf}
\IEEEoverridecommandlockouts
\usepackage{cite}
\usepackage{amsmath,amssymb,amsfonts}
\usepackage{algorithm}
\usepackage{algpseudocode}
\usepackage{graphicx}
\usepackage{comment}
\usepackage{url}
\usepackage{hyperref}
\usepackage{textcomp}
\usepackage{xcolor}
\usepackage[nolist]{acronym}
\usepackage{tikz}
\usepackage{siunitx}
\usetikzlibrary{shapes.geometric, arrows, positioning}

\tikzstyle{block} = [rectangle, rounded corners, minimum width=3cm, minimum height=1cm,text centered, draw=black, fill=blue!20]
\tikzstyle{input} = [ellipse, minimum width=2cm, minimum height=1cm, text centered, draw=black, fill=green!20]
\tikzstyle{output} = [ellipse, minimum width=2cm, minimum height=1cm, text centered, draw=black, fill=red!20]
\tikzstyle{arrow} = [thick,->,>=stealth]

\def\BibTeX{{\rm B\kern-.05em{\sc i\kern-.025em b}\kern-.08em
    T\kern-.1667em\lower.7ex\hbox{E}\kern-.125emX}}
\begin{document}

\title{Online Velocity Profile Generation and Tracking for Sampling-Based Local Planning Algorithms in Autonomous Racing Environments}
\author{Alexander Langmann, Levent Ögretmen, Frederik Werner and Johannes Betz 
\thanks{A. Langmann, J. Betz are with the Professorship of Autonomous Vehicle Systems, TUM School of Engineering and Design, Technical University of Munich, 85748 Garching, Germany; Munich Institute of Robotics and Machine Intelligence (MIRMI), corresponding author: 
\href{mailto:alexander.langmann@tum.de}{alexander.langmann@tum.de}}
\thanks{L. Ögretmen is with the Chair of Automatic Control, Technical University of Munich, 85748 Garching, Germany (e-mail: \href{mailto:levent.oegretmen@tum.de}{levent.oegretmen@tum.de})}
\thanks{F. Werner is with the Institute of Automotive Technology, TUM School of Engineering and Design, Technical University of Munich, 85748 Garching, Germany; Munich Institute of Robotics and Machine Intelligence (MIRMI)}
}

\vspace*{-2cm}
\noindent
\begin{center}
\fbox{
  \begin{minipage}{0.95\textwidth}
    \footnotesize
    © 2025 IEEE. Personal use of this material is permitted. Permission from IEEE must be obtained for all other uses, in any current or future media, including reprinting/republishing this material for advertising or promotional purposes, creating new collective works, for resale or redistribution to servers or lists, or reuse of any copyrighted component of this work in other works.
  \end{minipage}
}
\end{center}
\vspace{1cm}

\maketitle
\begin{abstract}
This work presents an online velocity planner for autonomous racing that adapts to changing dynamic constraints, such as grip variations from tire temperature changes and rubber accumulation. The method combines a forward-backward solver for online velocity optimization with a novel spatial sampling strategy for local trajectory planning, utilizing a three-dimensional track representation. The computed velocity profile serves as a reference for the local planner, ensuring adaptability to environmental and vehicle dynamics. We demonstrate the approach’s robust performance and computational efficiency in racing scenarios and discuss its limitations, including sensitivity to deviations from the predefined racing line and high jerk characteristics of the velocity profile.
\end{abstract}

\section{Introduction}
\subsection{Motivation}
High-performance motion planning and control for autonomous systems has garnered significant attention when it comes to investigating the capabilities of autonomous vehicles in terms of high-speed obstacle evasion and collision avoidance. Autonomous racing, as a highly dynamic domain, can serve as an ideal testing environment to push the boundaries of these technologies~\cite{betz2023TUM}. A critical challenge in this context is to ensure that an autonomous race car can quickly adapt to changing friction conditions~\cite{Werner.2024}. 
Relying on offline-generated velocity profiles, such as those derived from an offline optimized trajectory~\cite{brayshaw2005quasi, veneri2020freetrajectory, lovato2022threedimensional, rowold2023Online, perantoni2015Optimal}, introduces significant limitations. These profiles are static and do not adapt to changing dynamic environments. In reality, track conditions change due to factors such as varying tire temperature and rubber deposition accumulation, which consequently affect the feasible speed profile. 
One promising approach to address this challenge is the online adaptation of the velocity profile. This study presents an online velocity profile generation method based on an offline-computed, fixed path that adjusts to varying dynamic constraints in a three-dimensional representation of the race track. We further present a trajectory generation strategy in the spatial domain for sampling-based local planners to accurately represent apex locations and geometric properties of the race track. This strategy utilizes the online velocity profile to maximize performance in competitive single- and multi-vehicle scenarios. Additionally, we provide a discussion on the shortcomings and potential future improvements of our approach.

\begin{figure}
    \centering
    \includegraphics[width=1.0\linewidth]{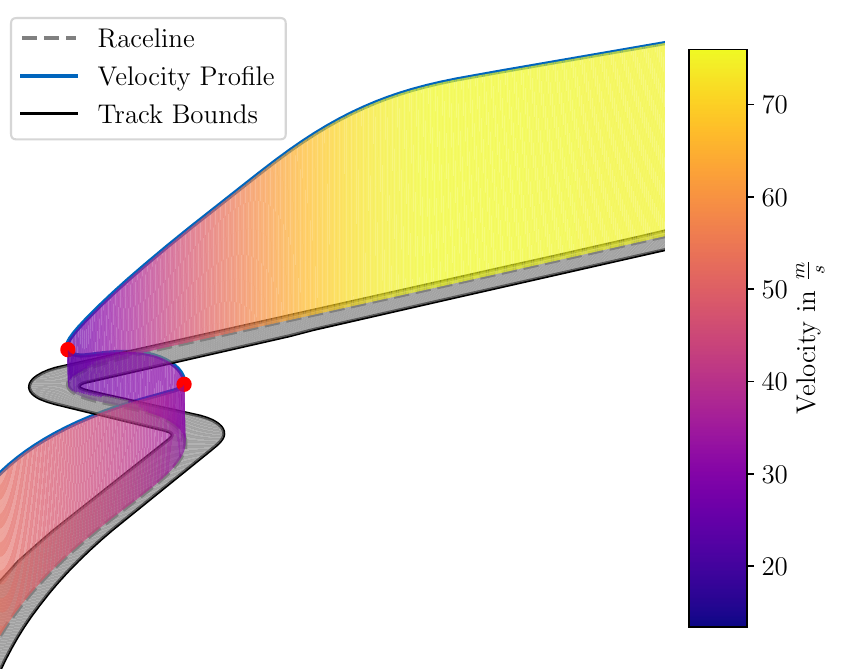}
    \caption{An exemplary speed profile of a race car for turns 6 and 7 at Yas Marina Circuit, Abu Dhabi. Two apexes (red dots) are identified on the depicted section of the track.}
    \label{fig:3D_vel}
\end{figure}

\subsection{Related Work}
Betz et al. \cite{betz2022Autonomous} provide a comprehensive review of motion planning techniques for autonomous racing, categorizing them into global and local planning. Global planning focuses on generating a time-minimal trajectory over the entire race track or a portion of the track and can be performed either offline or online. Heilmeier et al. \cite{heilmeier2020Minimum} formulate the problem as a quadratic optimization task to compute a minimum-curvature path and use a \ac{FW-BW} solver to obtain a minimum-time velocity profile. Kapania et al. \cite{kapania2016Sequential} adopt a two-step iterative approach, first optimizing the velocity profile along a fixed path and then refining the path based on this profile until convergence. Christ et al. \cite{christ2021Timeoptimal} incorporate a double-track vehicle model to account for varying road-tire friction potential. In addition to time-optimality, Ebbesen et al. \cite{ebbesen2018Timeoptimal} integrate an optimal energy management strategy for Formula E racing while minimizing lap time.

Beyond two-dimensional track representations \cite{veneri2020freetrajectory}, recent studies have addressed minimum-lap-time optimization in three-dimensional environments \cite{lovato2022threedimensional}. Rowold et al. \cite{rowold2023Online} propose an online race line optimization method that adapts to upcoming track segments in three dimensions but does not account for dynamic obstacles. Subosits et al. \cite{subosits2019Racetrack} introduce an adaptive global trajectory replanning approach, allowing a modification of the race line when encountering obstacles or deviating significantly from the reference path.

When the path of the race line is known, the velocity profile can be optimized to meet specific objectives. Lipp et al. \cite{lippMinimumTime} present a generalized framework for velocity optimization in robotic applications, while Lenzo et al. \cite{lenzo2020Simple} propose a simple curve-parameterized approach as an alternative to track-progress-based formulations. Herrmann et al. \cite{herrmann2021RealTime} introduce an optimization-based velocity planner that accounts for friction potential and energy constraints. Subotits et al. \cite{subosits2015Autonomous} generate a velocity profile with a \ac{FW-BW} solver on a topology to improve driving in hilly areas, but relying on the velocity profile of an offline optimization rather than online apex detection. Piccinini et al. \cite{piccinini2024Computationally} employ a neural network to generate motion primitives and compute feasible velocity profiles using a \ac{FW-BW} solver in two dimensions. Patton \cite{Patton2013} proposes an offline method for a fixed path by first constructing a boundary speed profile based on a purely lateral dynamic constraint. He then identifies \emph{critical points} and applies an \ac{FW-BW} solver to derive a feasible limit speed profile that respects combined lateral, longitudinal and yaw acceleration limits.

Local planning refines the globally planned race line by generating short-horizon trajectories that adapt to dynamic conditions. Many local planners utilize sampling-based methods, both in road traffic \cite{trauth2024FRENETIX, werling2010Optimal, wursching2021SamplingBased} and in racing scenarios \cite{raji2022Motiona, ogretmen2024SamplingBaseda, ogretmen2022Smooth, piazza2024MPTreea}. Also, hybrid techniques that combine graph- and sampling-based strategies \cite{ogretmen*2023Hybrida} are employed. These approaches ensure real-time feasibility but typically assume a static velocity profile along the reference line. Furthermore, the local trajectories are typically generated in a temporal domain, resulting in a spatial mismatch of, e.g., the braking point when deviating from the race line speed profile, typically due to overtaking maneuvers.

In summary, existing research offers a variety of techniques for generating and adapting race lines, as well as optimizing velocity profiles. However, most approaches either optimize velocity profiles offline or fail to integrate real-time velocity adjustments within a three-dimensional track environment. Furthermore, few local planning methods incorporate online velocity adaptation in combination with obstacle avoidance, and existing methods do not provide a way to follow the online velocity profile while ensuring that the spatial characteristics of the track are respected.

\subsection{Contribution}
This paper addresses the identified gap in the research landscape by proposing an approach that dynamically recomputes the velocity profile of the race line, based on varying vehicle dynamic constraints while complying with the vehicle's acceleration limits in a three-dimensional racing environment. We also introduce a spatial sampling technique to accurately incorporate the geometric properties of the race track, such as the location of an apex, into the local trajectories. We present experiments discussing how the online-generated velocity profile compares against an offline optimized velocity profile when utilized as a reference for a local sampling-based planning approach in single- and multi-vehicle scenarios.
Our work delivers three key contributions:

\begin{itemize}
	\item We provide a real-time capable calculation of a velocity profile on a three-dimensional race track based on a fixed path, capable of continuously adapting to varying vehicle dynamic constraints.
	\item We introduce a novel trajectory generation strategy for sampling-based planners that operates in the spatial domain. This approach ensures that the generated trajectories accurately reflect the longitudinal characteristics of the track and the velocity profile by aligning trajectory samples with key spatial features such as apex locations.
	\item We demonstrate our approach's effectiveness in single- and multi-vehicle autonomous racing scenarios in simulation and provide a discussion on the identified strengths and weaknesses.
\end{itemize}

\section{Methodology}
\begin{figure}
    \centering
    \includegraphics[width=1.0\linewidth]{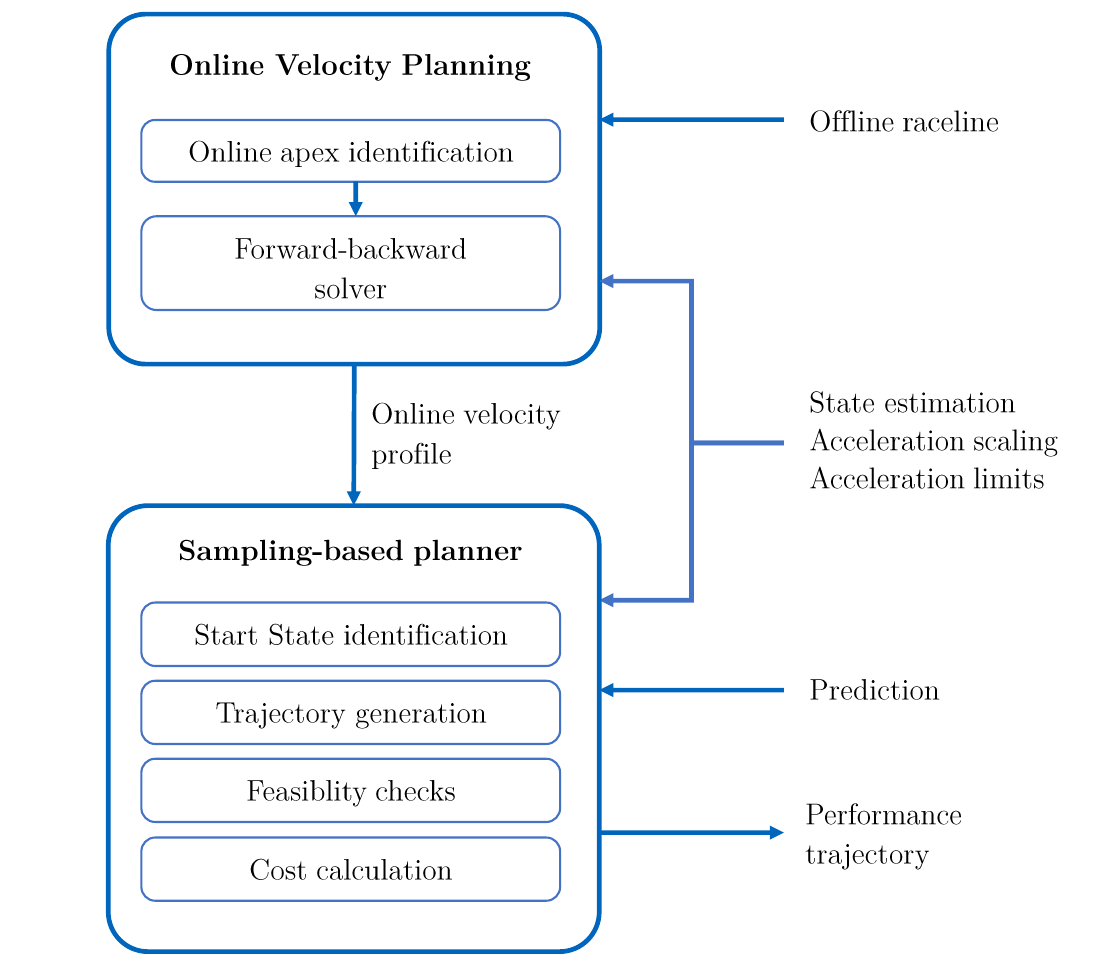}
    \caption{An overview of the proposed online velocity planning framework.}
    \label{fig:schaubild}
\end{figure}

This section presents the online velocity profile generation, its integration into a sampling-based trajectory planner, and the spatial-domain trajectory generation strategy. An outline of our framework is shown in Figure \ref{fig:schaubild}.

\subsection{Three-dimensional Track Representation}
\label{subsec:3d_track_rep}
We use the three-dimensional track representation from \cite{perantoni2015Optimal} to describe the drivable surface. It utilizes a parametrized curve 
\begin{equation}
    \mathcal{C} = \{ \mathbf{c}(s) \in \mathbb{R}^3 | s \in [0, s_{\mathrm{lap}}]\}
\end{equation}
as the spine of the drivable surface, which consists of concatenated flat planes. The orientation of the flat planes is described by Euler angles in the zyx convention $\phi(s), \mu(s)$ and $\theta(s)$. Since race tracks form a closed loop, $\mathbf{c}(s=0) = \mathbf{c}(s=s_{\mathrm{lap}})$ holds, where $s_{\mathrm{lap}}$ is the length of the spine. As the spine, also called reference line, we use an offline generated, three-dimensional race line based on \cite{rowold2023Online}. A coordinate system moves with the reference line $\mathcal{C}$, which we consider as the road frame. Following the formulae from Frenet-Serret, the road frame is orthonormal, consisting of a tangent vector $\mathbf{t}(s)$, a normal vector $\mathbf{n}(s)$ that lies on the road plane and a bi-normal vector $\mathbf{b}(s) = \mathbf{t}(s) \times \mathbf{n}(s)$. A point on the drivable surface is described by the progress $s$ along the reference line and its lateral offset from the reference line $n$.
As shown in \cite{rowold2023Online}, the angular velocity $\mathbf{\Omega}$ of the road frame with respect to the progress $s$ expressed in the road frame is 
\begin{equation}
\mathbf{\Omega} = 
\begin{bmatrix}
\Omega_{\mathrm{x}} \\
\Omega_{\mathrm{y}} \\
\Omega_{\mathrm{z}}
\end{bmatrix}
=
\begin{bmatrix}
1 & 0 & -s_{\mu} \\
0 & c_{\phi} & c_{\mu} s_{\phi} \\
0 & -s_{\phi} & c_{\mu} c_{\phi}
\end{bmatrix}
\cdot
\begin{bmatrix}
\phi' \\
\mu' \\
\theta'
\end{bmatrix},
\end{equation}
where $s_{\square}$ and $c_{\square}$ denote $\sin(\square)$ and $\cos(\square)$ and $\square'$ denotes the derivative with respect to the progress $s$. For the remainder of this paper, the term \textit{curvature} is used equivalently to $\Omega_{\mathrm{z}}$.

\subsection{Vehicle Dynamic Constraints}
\label{subsec:acceleration_constraints}
\begin{figure}
    \centering
    \includegraphics[width=1.0\linewidth]{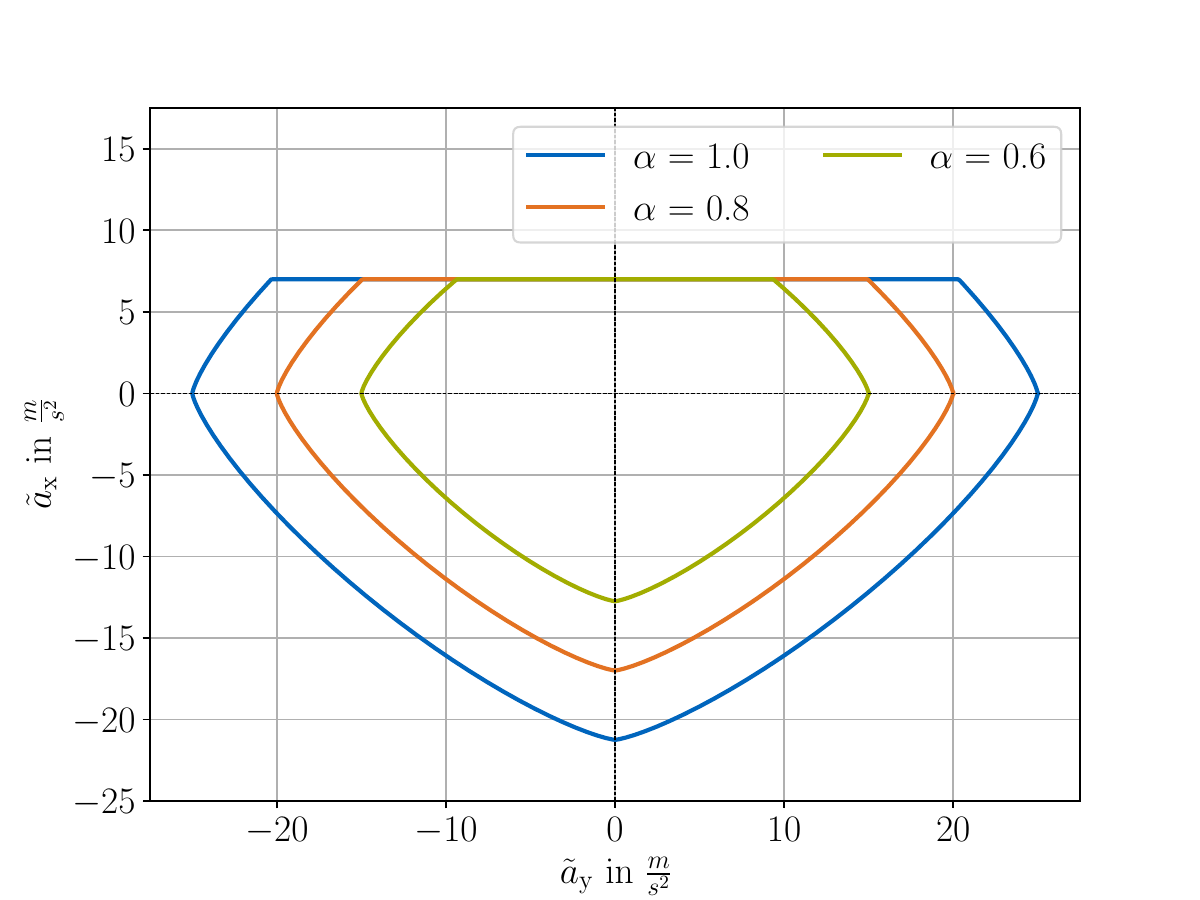}
    \caption{Exemplary gg-diagrams for $v = $ \SI{40}{m/s}, $\tilde{g} = $ \SI{9.81}{m/s^{2}} and $\rho = 1.3$, adjusted by three different scaling factors $\alpha$. Note that the longitudinal acceleration potential $\tilde{a}_{\mathrm{x, eng}}$ is not affected by $\alpha$.}
    \label{fig:gggv}
\end{figure}

We model the vehicle dynamics using a simplified point-mass representation under a quasi-steady-state assumption, neglecting transient effects. The accelerations of the point-mass $\hat{a}_{x}, \hat{a}_{y}$ are assumed to be in the velocity frame $\mathcal{V}$, which corresponds to a rotation of the road frame around the unit vector $\mathbf{b}(s)$ by an angle $\hat{\chi}$ so that $\mathbf{t}(s)$ aligns with the velocity vector $\mathbf{v}$, i.e. the lateral component of $\mathbf{v}$ is zero. To transform the accelerations from the velocity frame to apparent accelerations, we use the relations
\begin{equation}
\label{eq:apparent}
\begin{bmatrix}
\tilde{a}_{\mathrm{x}} \\
\tilde{a}_{\mathrm{y}} \\
\tilde{a}_{\mathrm{z}}
\end{bmatrix}
=
\begin{bmatrix}
\hat{a}_{x} + g(c_{\mu} \, s_{\phi}\, s_{\hat{\chi}} - s_{\mu} \, c_{\hat{\chi}})\\
\hat{a}_{y} + g(s_{\mu} \, s_{\hat{\chi}} + c_{\mu} \, s_{\phi} \, c_{\hat{\chi}})\\
\hat{w}_{y} \, v + g(c_{\mu} \, c_{\phi})
\end{bmatrix}
\end{equation}
that already neglect insignificant terms. A derivation from \eqref{eq:apparent} is given in the appendix of \cite{rowold2023Online}.
For modeling the acceleration constraints of the vehicle, we use implicitly defined gg-diagrams that are extended by the apparent gravitational acceleration $\tilde{g}$ and depend on the velocity $v$, based on \cite{lovato2022threedimensional}. An example is shown in Figure \ref{fig:gggv}. The acceleration limits $\tilde{a}_{\mathrm{x, min}}$, $\tilde{a}_{\mathrm{y, max}}$ and $\tilde{a}_{\mathrm{x, max}}$ are the apparent acceleration limits, i.e. the accelerations that would be measured by the vehicle's \ac{IMU}. They describe the vertices of the diagrams mainly due to the tire limits of the vehicle. The positive longitudinal limit $\tilde{a}_{\mathrm{x, eng}}$ takes limited engine power and aerodynamic drag of the vehicle into account and cuts off the diagram in the positive section. We use the exponent $\rho$ based on \cite{rowold2023Online} to morph the shape of the gg-diagram. To account for varying acceleration limits, we introduce a scaling factor $\alpha \in (0, 1]$ that scales the acceleration limits of the diamond linearly. If the inequalities
\begin{equation}
\begin{aligned}
    \tilde{a}_{\mathrm{x}} &\leq \tilde{a}_{\mathrm{x, eng}} \text{ and}\\
    1 &\leq \left(\frac{\tilde{a}_{\mathrm{x}}}{\alpha \, \tilde{a}_{\mathrm{x, lim}}}\right)^{\rho} + \left(\frac{\tilde{a}_{\mathrm{y}}}{\alpha \, \tilde{a}_{\mathrm{y, max}}}\right)^{\rho}\\
    \text{with } \tilde{a}_{\mathrm{x, lim}} &=
    \begin{cases}
    \tilde{a}_{\mathrm{x, max}} & \text{ if } \tilde{a}_{\mathrm{x}} > 0,\\
    \tilde{a}_{\mathrm{x, min}} & \text{ else,}
\end{cases}
\end{aligned}
\end{equation}
hold, an acceleration state of the vehicle is considered feasible.

\subsection{Online Apex Detection}
\label{subsec:apex_detection}
We define an apex as the location of a local minimum in the feasible velocity profile along the race track. This location does not necessarily coincide with a local maximum of the curvature $\Omega_{\mathrm{z, max}}$ along the race line in three-dimensional environments and is subject to change with varying grip due to $\tilde{g}$ depending on $v$. To identify the apexes in the interval $s_{\mathrm{ahead}} \in [s_{\mathrm{start}}, s_{\mathrm{start}} + h_{\mathrm{opt}}]$ ahead of the vehicle, apex candidate locations $s_{\mathrm{cand}, i}$ are identified by detecting local maxima of $\Omega_{\mathrm{z, max}}$. Calculating the admissible velocity at each point in the interval $s_{\mathrm{search}, i} \in [s_{\mathrm{cand}, i} - l/2, s_{\mathrm{cand}, i} + l / 2]$, where $l$ is the length of the interval, using fixed-point iteration yields the precise apex location. The initial velocity guess for a candidate point
 \begin{equation}
 \label{eq:v_guess}
    V_{\text{guess, cand}} = \sqrt{\alpha} \, V_{\text{off}}
\end{equation}
is derived from the velocity of the offline race line $V_{\text{off}}$, adjusted to the current acceleration limits under the assumption of the absence of longitudinal acceleration. An explanation to \eqref{eq:v_guess} is given in the appendix.
For each apex candidate, the maximum admissible velocity is refined iteratively using the relation
\begin{equation}
    V_{\text{new}} = \sqrt{\frac{\hat{a}_{\mathrm{y}}}{\Omega_{\mathrm{z}}}}.
\end{equation}
The admissible velocity is found if the algorithm terminates, i.e. $|V_{\text{new}} - V_{\text{old}}| < \epsilon$ is fulfilled, where $\epsilon$ is a predefined tolerance. If the iteration does not stop within a maximum number of steps, $V_{\text{guess, cand}}$ is assumed as the admissible velocity.
The apex location $s_{\mathrm{apex, i}}$ is identified on $s_{\mathrm{search}, i}$ where the admissible velocity is minimal. This procedure is performed for each apex candidate location and the corresponding search interval in the upcoming track section.

\subsection{Velocity Profile Generation}
\label{subsec:vel_profile_gen}
We divide the upcoming track section that contains $m$ apexes into $m+1$ segments at the apex locations. For each segment except the furthest away from the current vehicle location, we use a \ac{FW-BW} solver to analytically calculate the feasible velocity profile for the corresponding segment based on the path of a race line generated offline. In the last segment, if not ended by an apex location, we only perform a forward integration step as no deceleration towards an apex is required.
Using the gg-diagrams from \eqref{subsec:acceleration_constraints}, the forward step for a discretization point $i$ on the race line calculates the longitudinal apparent acceleration potential 
\begin{equation}
\label{eq:ax_forw}
    \tilde{a}_{\mathrm{x, fw}} =  \min\left\{\tilde{a}_{\mathrm{x, eng}}, \tilde{a}_{\mathrm{x, max}} \, \left(1 - \left(\frac{\tilde{a}_{\mathrm{y}}}{\alpha \, \tilde{a}_{\mathrm{y, max}}}\right)^{\rho}\right)^{\frac{1}{\rho}}\right\}
\end{equation}
based on the current lateral apparent acceleration $\tilde{a}_{\mathrm{y}}$. The indices in \eqref{eq:ax_forw} are left out for brevity.
To obtain the longitudinal acceleration potential $\hat{a}_{\mathrm{x}}$ in the velocity frame $\mathcal{V}$, the transformation in \eqref{eq:apparent} is used.
The feasible velocity $V_{i+1}$ in the next discretization point 
\begin{equation}
\label{eq:update}
\begin{aligned}
V_{i+1} &= 
\begin{cases}
    V_{\mathrm{next}} & \text{if } V_{\mathrm{next}} \leq V_{\mathrm{max}}, \\
    V_{\mathrm{max}} & \text{else, } 
\end{cases}\\
\text{with } V_{\mathrm{next}} &= \sqrt{\max\{V_{i}^{2} + 2 \, \hat{a}_{\mathrm{x, fw}} \, \Delta s, 0\}}
\end{aligned}
\end{equation}
is obtained assuming constant acceleration on the interval $\Delta s$ between two discretization points. It further considers a potentially imposed speed limit $V_{\mathrm{max}}$ in a racing situation. If $V_{\mathrm{next}} > V_{\mathrm{max}}$ holds, the acceleration at $s_{i}$ is corrected by solving \eqref{eq:update} with $V_{i+1} = V_{\mathrm{max}}$ for $\hat{a}_{\mathrm{x, fw}}$.
The backward step does not consider the longitudinal machine limits $\tilde{a}_{\mathrm{x, eng}}$, which yields the feasible longitudinal deceleration
\begin{equation}
    \tilde{a}_{\mathrm{x, bw}} = \tilde{a}_{\mathrm{x, min}} \, \left(1 - \left(\frac{\tilde{a}_{\mathrm{y}}}{\alpha \, \tilde{a}_{\mathrm{y, max}}}\right)^{\rho}\right)^{\frac{1}{\rho}}.
\end{equation}
The feasible velocity is obtained using an analogous procedure as the forward step from \eqref{eq:update}. Once the forward and backward velocity vectors at the discretized points of the race line $\mathbf{V}_{\mathrm{fw}}$ and $\mathbf{V}_{\mathrm{bw}}$ are generated, the overall feasible velocity profile $\mathbf{V}_{\mathrm{feas}}$ is obtained by
\begin{equation}
    \mathbf{V}_{\mathrm{feas}} = \min \left\{\mathbf{V}_{\mathrm{fw}}, \mathbf{V}_{\mathrm{bw}}\right\}
\end{equation}
and the corresponding accelerations are taken from the minimal velocity profiles. Finally, the time stamps of the race line are updated, taking into account the calculated online velocity profile.

\subsection{Sampling-based Local Trajectory Planner}
\label{subsec:sp_planner}
For the local trajectory planning task, we utilize a sampling-based trajectory planning approach based on \cite{ogretmen2024SamplingBaseda}. It utilizes the race line with the updated velocity profile from Section \ref{subsec:vel_profile_gen} as its reference profile. 

The start state of a trajectory planning step is determined by matching the current progress along the reference line $s$, received by a state estimation module, on the trajectory planned in the previous cycle. From this matching point, the previously planned trajectory is traversed by a predefined time $t_{\mathrm{const}}$ to compensate the calculation time required by the motion planning step. If no previous trajectory is available, e.g. in case of the initial planning step, the vehicle state obtained from the state estimation serves as the start state. 

We base our trajectory generation step on \cite{werling2010Optimal, ogretmen2024SamplingBaseda} to generate purely jerk-optimal and relative trajectories that have a longitudinal tendency towards the velocity profile of the race line.
While these approaches perform trajectory generation over a fixed temporal horizon $T$, we extend the sampling strategy to generate relative trajectories with a fixed spatial horizon $S$ to capture brake points adequately when off the velocity profile, e.g., when performing an overtake. We require the derivatives
\begin{equation}
\label{eq:domain_shift}
    \ddot{s}(s)=\frac{\ddot{s}(t)}{\dot{s}(t)}, n'(s)=\frac{\dot{n}(t)}{\dot{s}(t)} \text{ and } n''(s)=\frac{\ddot{n}(t) - n'\ddot{s}(t)}{\dot{s}(t)^{2}}
\end{equation}
to perform the shift of the start and end conditions $[s(t), \dot{s}, \ddot{s}, n(t), \dot{n}, \ddot{n}] \xrightarrow{} [s(s), \dot{s}(s), \ddot{s}(s), n(s), n', n'']$ to the spatial domain. Since the length of the trajectory is fixed by $S$, a third-order polynomial is sufficient for the longitudinal sampling step in the spatial mode instead of a fourth-order polynomial in the temporal mode. Once generated, the trajectories are transformed back to the temporal domain using the relations from \eqref{eq:domain_shift} to remain compatible with the temporal generation strategies and the cost functional. We cut trajectories to the temporal horizon $T$ if they exceed it.

The generated trajectories are checked for feasibility. We perform three distinct checks to ensure a trajectory is drivable. First, the curvature of the trajectory is checked so it remains below the maximum curvature $\kappa_{\mathrm{max}} = \frac{1}{r}$ where $r$ denotes the minimum turning radius of the vehicle. Second, we ensure every trajectory point remains fully on the race track to avoid unintentionally leaving the track. Third, the acceleration of the trajectory is checked to remain within the gg-diagrams. Here, we allow slight violations so that trajectories that exceed the limits by a small margin are not sorted out immediately, but costs are assigned in \eqref{eq:costs}.

To evaluate the trajectories in terms of suitability to the driving task within the planning horizon $T$, we evaluate a cost functional 
\begin{equation}
\label{eq:costs}
    C = \sum_{i=0}^{N}\int_{0}^{T}w_{i} c_{i}(t) \space dt
\end{equation}
for each trajectory that penalizes undesired behavior. In total, we calculate six distinct cost terms $w_{i}c_{i}$: lateral deviation from the race line, deviation from the race line curvature, deviation from the target velocity, risk and severity of a collision, and violations of the acceleration limits. The weights $w_{i}$ are tuned to produce favorable behavior in racing situations. 

\section{Results}
Our experiments are conducted in simulation on Yas Marina Circuit in Abu Dhabi and consist of single- and multi-vehicle scenarios. We focus on the track section of turns 6 and 7 as these turns are difficult to traverse at the performance limit, including high and slow speed sections with high curvatures. The used vehicle platform corresponds to a Super Formula chassis with modifications to be able to run autonomously as used in the \ac{A2RL}. To ensure determinism of our experiments, we fix the elapsed time for a planning step. We neglect potential influences by inaccuracies in the state estimation, control or perception module by ensuring the planned trajectory is tracked perfectly and the current and future positions of opponent vehicles are exactly known. All simulation runs are performed on an Intel Core i7-1270P CPU. The relevant simulation parameters are summarized in Table \ref{tab:params}.

\begin{table}[!htbp]
    \centering
    \caption{Simulation parameters}
    \begin{tabular}{ccc}
        \hline
        Parameters & Value & Unit\\
        \hline
        Simulation time per planning step & 100 & ms\\
        Temporal horizon of the sampling planner $T$& 4& s\\
        Velocity optimization horizon $h_{\mathrm{opt}}$ & 600 & m\\
        Race line discretization $\Delta s$& 1.0 & m
    \end{tabular}
    \label{tab:params}
\end{table}

\subsection{Longitudinal Tracking of the Offline Race Line}
\begin{figure}[!htbp]
    \centering
    \includegraphics[width=1.0\linewidth]{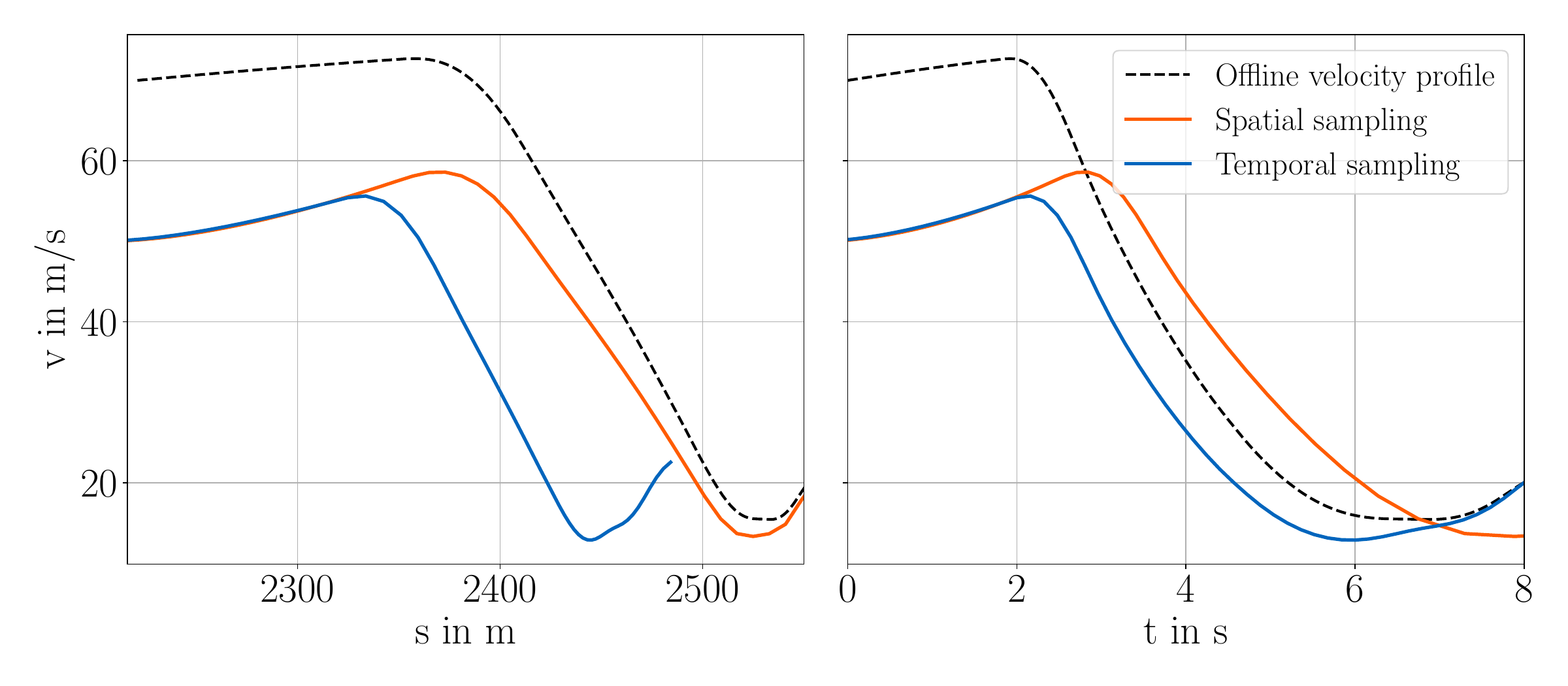}
    \caption{Comparison of relative longitudinal sampling strategies in the spatial and temporal domain in turns 6 and 7.}
    \label{fig:sampling_comp}
\end{figure}

In the first experiment, we perform the temporal and spatial velocity profile generation from Section \ref{subsec:sp_planner} at the location $s=$ \SI{2200}{m} and investigate the velocity tracking capabilities when following an offline generated velocity profile. Since it is shown in \cite{ogretmen2024SamplingBaseda} that perfect tracking is possible if the start state lies on the velocity profile, we create a difficult scenario by setting the start velocity approx. \SI{20}{m/s} below the offline velocity profile. Figure \ref{fig:sampling_comp} shows the planned velocity profiles in the spatial and temporal domain. When transforming the temporally relative profile from time- to space-domain, the planned profile decelerates unnecessarily early and accelerates well before the apex location of the track is reached, not matching the geometric properties of the turns.
The spatially relative trajectory, not capturing the race line's characteristics with respect to time, accurately incorporates the location of the brake point and apex into its velocity profile. We deduce that spatially sampled trajectories relative to the race line fulfill the tracking task of the race line better than temporal trajectories and use spatially sampled trajectories for the remainder of our experiments.

\subsection{Reduced Acceleration Limits}
\label{subsec:red_grip}
\begin{figure}[!htbp]
    \centering
    \includegraphics[width=1.00\linewidth]{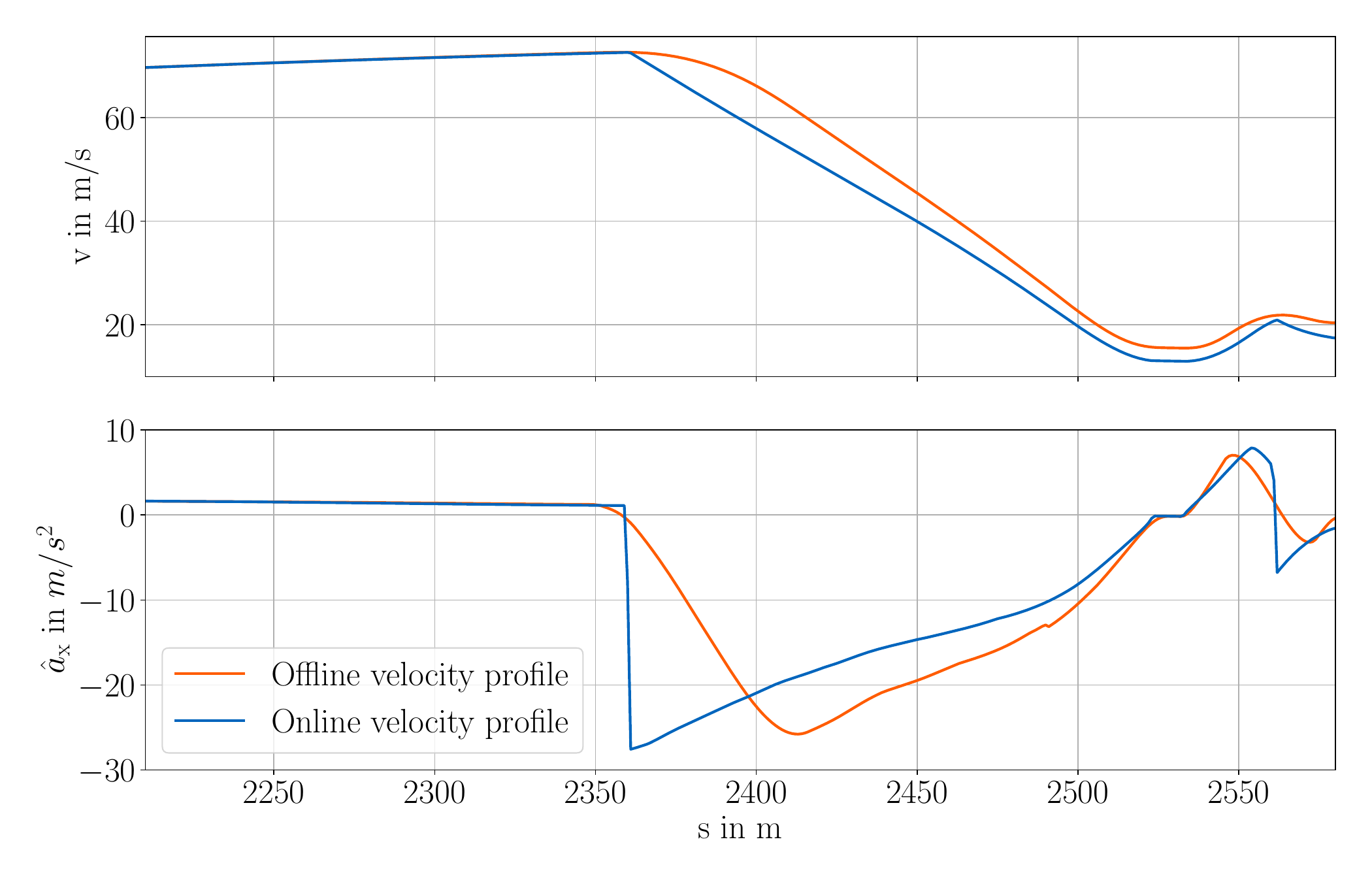}
    \caption{The online velocity profile respecting an acceleration limit scaling of $\alpha=0.7$ and the offline generated profile. The lower plot shows the corresponding accelerations.}
    \label{fig:ovp_comp}
\end{figure}

We model the track section from $s=$ \SI{2000}{m} to $s=$ \SI{2600}{m} with reduced grip to mimic a real-time grip update and model this by an acceleration limits scaling factor of $\alpha = 0.7$. We compare the resulting online velocity profile to an offline velocity profile that is uninformed of the reduction in grip. This comparison is shown in Figure \ref{fig:ovp_comp}.
The apex speed at $s=$ \SI{2520}{m} is lower in the online velocity profile, incorporating the reduced acceleration limits. In section where $\hat{a}_{\mathrm{x}}$ is low, the $\hat{a}_{\mathrm{x}}$ curves of both profiles are qualitatively similar. However, in sections with sudden transitions between high positive and negative accelerations, the unregularized nature in regards to transient state changes of the \ac{FW-BW} solver becomes evident, resulting in instantaneous changes in acceleration. These abrupt transitions may not be ideal for real-world execution due to physical actuator limitations. 

\begin{figure*}[!htbp]
    \centering
    \includegraphics[width=0.75\textwidth]{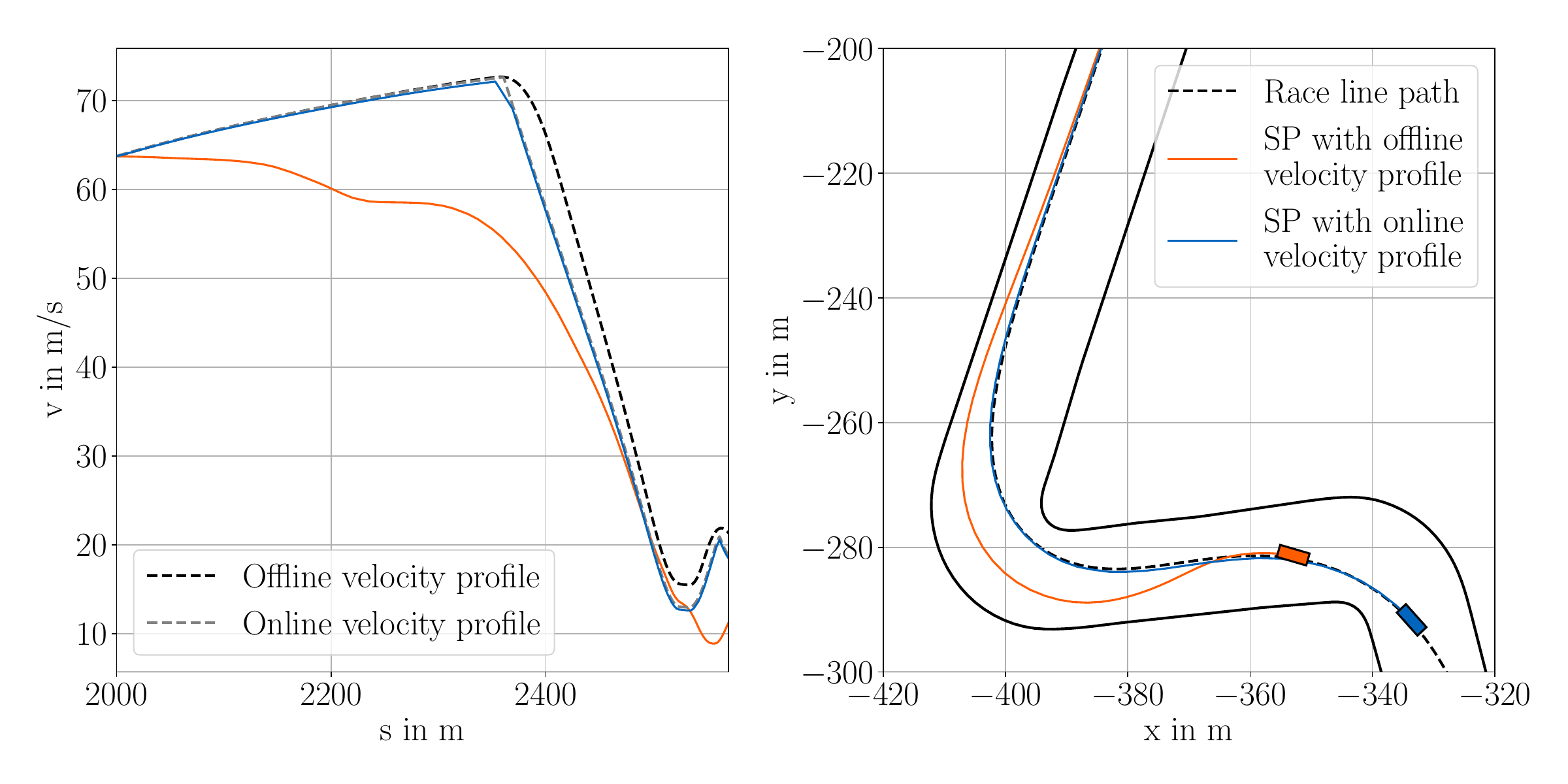}
    \caption{Velocity profiles (left) and paths (right) produced by the sampling planner (SP) when tracking the online and offline velocity profile. The rectangles indicate the vehicle position at the time frame when the sampling planner using the online velocity profile is at $s=$\SI{2600}{m}.}
    \label{fig:sp_comp}
\end{figure*}

An investigation of the sampling-based planner tracking performance of the velocity profiles using the identical vehicle dynamic constraint set scaled by $\alpha$ is shown in Figure \ref{fig:sp_comp}. When using the offline race line as a reference, the sampling planner fails to reliably generate feasible trajectories. This results in an early braking maneuver and heavy deceleration after the apex at approx. $s=$ \SI{2520}{m}. The observed deviation from the reference path is a result of the attempt to limit vehicle dynamic constraint violation through an extension of the braking zone.
When using the online velocity profile, the sampling planner manages to adapt to the decreased grip conditions and follow both the race line velocity and the path while remaining feasible, resulting in a faster time by \SI{1.42}{s} for this track section as indicated in Table \ref{tab:times}. 

\subsection{Deviating From the Race Line}
\label{subsec:multi}
\begin{figure*}[!htbp]
    \centering
    \includegraphics[width=0.8\textwidth]{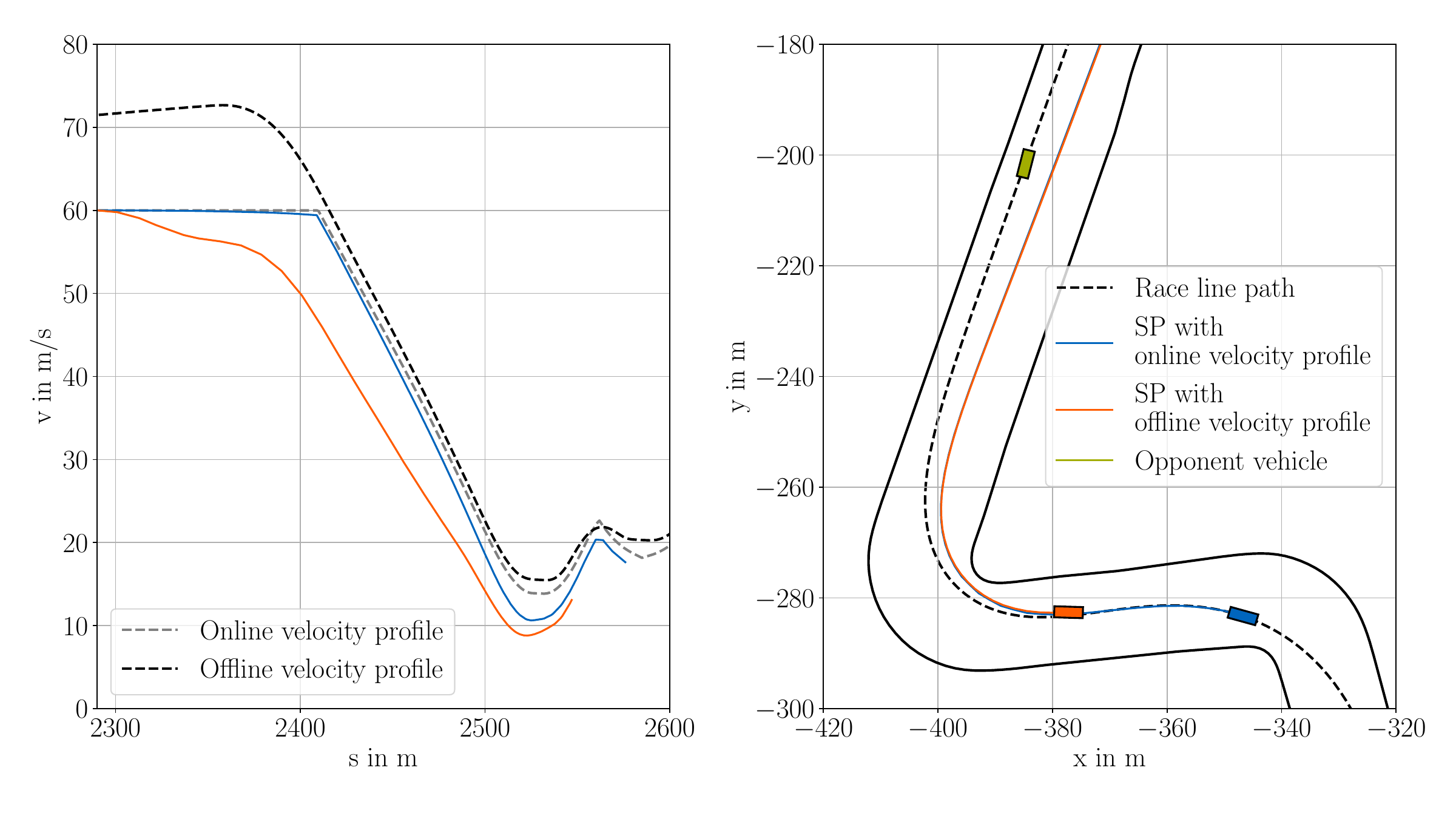}
    \caption{Velocity profiles (left) and paths (right) produced by the sampling planner (SP) with and without an online calculated velocity profile. The grip scaling factor was $\alpha$ set to 0.7. A static object was placed on the race line and the vehicle was initialized laterally off the race line path.}
    \label{fig:multi}
\end{figure*}

In a multi-vehicle racing situation, it is common to laterally deviate from the race line to perform an overtaking or yielding maneuver. To emulate this, a static object is placed on the race line at $s=$ \SI{2500}{m}, blocking the race line path at the entry to turn 6. The vehicle start state is at $s=$ \SI{2000}{m} with a lateral deviation of \SI{6}{m} to the left of the race line and the maximum allowed speed is \SI{60}{m/s}. To emulate reduced grip off the race line, we set $\alpha = 0.7$. Figure \ref{fig:multi} shows the resulting velocities and paths in this scenario.
Similarly to Section \ref{subsec:red_grip}, the sampling planner using the online velocity profile display closer velocity tracking compared to the sampling planner based on the offline velocity profile, generated with $\alpha=1.0$. Increased curvatures need to be achieved since the turn is negotiated on the inside of the race line, so the velocity profile cannot be tracked and a slower trajectory is chosen. The spatial sampling strategy enables the planner to adjust to the higher curvature by integrating the local velocity minimum at the apex location.
In terms of lateral tracking, using the offline velocity profile does not have an influence, as the paths of the online and offline velocity-based planners are almost identical. However, the velocity profile achieved with the offline race line is more cautious, resulting in a time difference of \SI{1.18}{s} for the shown track section.

\begin{table}[!htbp]
    \centering
    \caption{Sector time from s=2000m to s=2600m }
    \begin{tabular}{ccc}
        \hline
        Planner configuration & Time & 
        Difference\\
        \hline
        Sampling planner with online velocity profile & \SI{13.55}{s} & -\\
        Sampling planner with offline velocity profile & \SI{14.97}{s} & \SI{1.42}{s} \\
    \end{tabular}
    \label{tab:times}
\end{table}

\subsection{Runtime Analysis}
During our simulation runs, the runtime of the entire trajectory planning step, including online velocity profile generation, averaged at 114 ms. The online velocity profile generation itself achieved an average runtime of 43 ms using a Python implementation, demonstrating its feasibility for online application.
The run time increases linearly with the number of identified apexes within the velocity optimization horizon $h_{\mathrm{opt}}$ since every apex requires the \ac{FW-BW} calculation of an additional segment. Furthermore, the distance between discretized points $\Delta s$ of the race line path influences the runtime linearly. The simulation was executed on a single CPU core, but future implementations could offload the velocity profile calculation to a separate core to run asynchronously, improving overall computational performance.
\section{Discussion}
This section discusses the strengths and limitations of our proposed approach. We address the key benefits of our framework and highlight the limitations that need to be addressed in future work.

\subsection{Adaption to Changing Track Conditions}
Our proposed framework demonstrates the ability to adapt to changing vehicle dynamic constraints and calculates a dynamically feasible velocity profile for a given path. Offline computed velocity profiles, being static, fail to adapt to real-time changes in grip, leading to conservative or infeasible trajectories.
In terms of performance and lap time, we show that an adaption of the reference velocity profile to the online sampling-based planner based on the current vehicle dynamic constraints is essential and can result in time differences of over one second in an exemplary chicane section of a race track alone. However, reliance on accurate real-time data for grip estimation can be a potential limitation, therefore requiring accurate and reliable grip estimation to enhance performance in real-world tests.

\subsection{Robustness in Multi-Vehicle Situations}
In the online velocity profile generation, the resulting profile is guaranteed to be feasible only when traversing on the race line path. If lateral deviations occur, which is typical for multi-vehicle scenarios, the profile provides a good estimation of the feasible velocity profile but is not guaranteed to be feasible for the updated path. This poses a major limitation of our approach, making it suitable for single-vehicle racing, but only partially suitable for multi-vehicle scenarios. As shown in Section \ref{subsec:multi}, the spatial sampling strategy provides robustness to lateral deviations and enables the sampling planner to track the online velocity profile without acceleration limit violations when traveling off the racing line. We are aware that future research needs to address this by investigating real-time updates on the velocity profile for paths deviating from the initial race line path.

\subsection{Acceleration Constraints and Jerk Regularization}
The current framework models the vehicle dynamics using a point-mass model representation, providing a computationally efficient way to evaluate if accelerations in a state are feasible. However, it does not capture the full complexity of vehicle dynamics, particularly in situations where transient effects are non-negligible. These effects are not integrated into the online velocity profile, which might lead to unstable driving states while remaining feasible in the point-mass formulation, failing to capture safety critical effects. Works like Patton~\cite{Patton2013} offer potential solutions to respect transient vehicle limitations, even with quasi-steady-state point-mass model. Additionally, we want to investigate the use of a more sophisticated model to represent the vehicle dynamics and include handling transient effects in future work.

Another key limitation observed is the lack of jerk regularization in the online velocity profiles generated by the \ac{FW-BW} solver. The abrupt changes in acceleration can be problematic for real-world implementation due to physical actuator limitations. Incorporating jerk constraints into the velocity profile generation process could smooth out these transitions, making the profiles more feasible for real-world execution. This would enhance the overall drivability and could reduce stress on control algorithms. Additionally, smoother acceleration profiles could reduce wear on the vehicle's components. Future work could explore the integration of jerk constraints into the optimization process or evaluate smoothing techniques as a post-processing step.
\section{Conclusion and Future Work}
In this paper, we present a method for online velocity planning in autonomous racing, integrating a dynamically feasible velocity profile with a sampling-based local planner in a three-dimensional track representation. We further introduce a novel spatial sampling strategy for local planners. Our approach demonstrates the ability to adapt to changing track conditions in real-time, ensuring robust performance in both single- and multi-vehicle scenarios in a three-dimensional race track representation.

Future work will focus on several areas to further enhance the framework. First, incorporating more detailed vehicle dynamics models will improve the accuracy of the generated velocity profiles and trajectories. Second, improved techniques to handle driving laterally off the race line while maintaining feasibility shall be explored.
Third, methods to smooth the acceleration profile, such as incorporating jerk constraints, will be explored to ensure physical feasibility. Finally, extensive real-world validation will be conducted to test the framework's performance and robustness in actual racing environments.

Overall, the proposed framework provides a solid foundation for high-performance motion planning in autonomous racing, with significant potential for further improvements and real-world applications.

\bibliographystyle{IEEEtran}
\bibliography{literature.bib}

\begin{acronym}
\acro{AVs}{autonomous vehicles}
\acro{LVMS}{Las Vegas Motor Speedway}
\acro{IAC}{Indy Autonomous Challenge}
\acro{MPC}{Model Predictive Control}
\acro{A2RL}{Abu Dhabi Autonomous Racing League}
\acro{ODD}{operation design domain}
\acro{OCP}{optimal control problem}
\acro{RL}{reinforcement learning}
\acro{MARL}{multi-agent reinforcement learning}
\acro{IMU}{inertial measurement unit}
\acro{FW-BW}{forward-backward}
\end{acronym}


\appendix
\label{appendix}
Assuming no longitudinal acceleration present on an arbitrary point $i$ on the race line with curvature ${\Omega_{\mathrm{z}}}$ so that 
\begin{equation}
     V_{\text{i}}^{2} = \frac{\hat{a}_{\mathrm{y, i}}}{\Omega_{\mathrm{z}}}
\end{equation}
is valid and neglecting effects of three-dimensional nature, we use
\begin{equation}
\begin{aligned}
    \Omega_{\mathrm{z}} &= \frac{\hat{a}_{\mathrm{y,1}}}{V_{\text{1}}^{2}} = \frac{\hat{a}_{\mathrm{y,2}}}{V_{\text{2}}^{2}}\\ 
    &= \frac{\hat{a}_{\mathrm{y,1}}}{V_{\text{1}}^{2}} = 
    \frac{\alpha \cdot \hat{a}_{\mathrm{y,1}}}{V_{\text{2}}^{2}}
\end{aligned}
\end{equation}
to establish the relation between admissible velocities for varying acceleration limit scaling factors $\alpha$
\begin{equation}
    V_{2} = \sqrt{\alpha} \cdot V_{1}.
\end{equation}

\end{document}